\documentclass[11pt]{article}
\usepackage{color, colortbl}
\usepackage{paclic35}
\usepackage{times}
\usepackage{latexsym}
\usepackage{amssymb}
\usepackage{multirow}
\usepackage{subcaption,booktabs}
\usepackage{graphics,graphicx}
\usepackage{url}
\title{Under the Microscope: \\Interpreting Readability Assessment Models for Filipino}

\author{Joseph Marvin Imperial \\
  National University \\
  Manila, Philippines \\
  {\tt jrimperial@national-u.edu.ph} \\\And
  Ethel Ong \\
  De La Salle University \\
  Manila, Philippines \\
  {\tt ethel.ong@dlsu.edu.ph} \\}

\date{}

\begin{document}
\maketitle
\begin{abstract}
Readability assessment is the process of identifying the level of ease or difficulty of a certain piece of text for its intended audience. Approaches have evolved from the use of arithmetic formulas to more complex pattern-recognizing models trained using machine learning algorithms. While using these approaches provide competitive results, limited work is done on analyzing how linguistic variables affect model inference quantitatively. In this work, we dissect machine learning-based readability assessment models in Filipino by performing global and local model interpretation to understand the contributions of varying linguistic features and discuss its implications in the context of the Filipino language. Results show that using a model trained with top features from global interpretation obtained higher performance than the ones using features selected by Spearman correlation. Likewise, we also empirically observed local feature weight boundaries for discriminating reading difficulty at an extremely fine-grained level and their corresponding effects if values are perturbed.
  
\end{abstract}

\section{Introduction}

Readability assessment is the process of evaluating a certain piece of text or reading material in terms of reading difficulty. Likewise, reading difficulty can be expressed in various forms such as age level, grade level, or a number from a certain range defined by a book publisher \cite{deutsch-etal-2020-linguistic}. Through the years, this process has evolved from the use of handcrafted arithmetic formulas such as the Flesch-Kincaid Reading Ease \cite{kincaid1975derivation} and Dale-Chall \cite{dale1948formula} readability formulas to the use of supervised machine learning algorithms such as Logistic Regression and Support Vector Machines \cite{chatzipanagiotidis-etal-2021-broad,weiss-meurers-2018-modeling,xia-etal-2016-text,reynolds-2016-insights,vajjala-meurers-2012-improving}. Despite the significant growth in research history, several problems still pose as open challenges for the task such as the (a) availability of corpora and tools for linguistic feature extraction, (b) extrinsic evaluation, and (c) interpretation of linguistic predictors used which is arguably the most important of all \cite{vajjalatrends}. 


We describe the general readability assessment task as \textit{feature interpretation dependent} since identifying which optimal subset of linguistic features that can potentially impact readability levels is a strict and necessary part of the research process that should not be ignored. Recent works in readability analysis \cite{imperial2020exploring,hancke-etal-2012-readability} have used standalone, correlation-based feature selection techniques such as Spearman or Pearson correlation to get a better understanding of feature dependence and relationship. These methods, however, can be done even \textit{before} model training which may not be holistically predictive of features that trained machine learning models would eventually find useful \cite{kumar2018correlation}. In addition, these methods break down as they only measure the linear relationship of linguistic variables in contrast to a possibility of a non-linear relationship of features in the dataset.

In order to have a clear understanding of how a linguistic predictor affects model inference in readability assessment, the learned model weights of the said machine learning model should be extracted and analyzed \textit{after} training. In this way, one can survey and rank the features used by a model and that has contributed substantially towards obtaining the final output. Thus, we lay down our contributions for this study as follows:

\begin{enumerate}
    \item Feature selection through global interpretation of the state-of-the-art trained model in readability assessment for Filipino;
    \item Analysis of the performance of readability assessment models trained using top features from global interpretation against top features from a correlation method; and,
    \item Close-up analysis of top local features used by each grade level for readability analysis and their local weight boundaries through local surrogate interpretation.
\end{enumerate}

\section{Task Definition}
We define the readability assessment task as a supervised learning problem. Given a document or specifically, a reading material $d$, a feature vector $x = [x_{1},x_{2}\ldots,x_{n}]$ is extracted and a model $M$ is trained using said collection of linguistic predictors $X$ along with the gold label $Y$ or expert-identified readability level. From the trained model $M$, top $n$ features $g = [g_{1},g_{2} \ldots ,g_{n}]$ can be extracted through global interpretation and value boundaries $(a,b)$ where $a,b \in \mathbb{R}$ of said features $g_{i}$ can be identified through local surrogate interpretation.

\section{Readability Assessment Models for Filipino}
We survey different machine learning-based readability assessment models from previous works done for the Filipino language. We highlight the advantages and disadvantages of each model based on the data they used, scope of linguistic features, method of model training, and performance via select evaluation metrics. \linebreak 

\noindent\textbf{Guevarra \shortcite{Guevarra2011}.} This is the first ever work that utilized the supervised machine learning methodology for readability assessment in Filipino. This work used logistic regression as the primary algorithm for model development and used 7 linguistic properties spanning frequency of words, unique words, sentences, average syllable count, word occurrence, log of word frequency, and percentage of top words. For the data, 140 children's books from Adarna House\footnote{https://adarna.com.ph/}, the largest children's book publisher in the Philippines, were used. A relatively small error-rate was obtained using this approach during model evaluation with a score of 0.83 from the original grade level. \linebreak

\noindent\textbf{Bosque et al. \shortcite{bosquetagatris}.} This work pioneered the use of simple term frequencies and term-frequency inverse document frequency (TF-IDF), an NLP-based feature which represents each instance of the corpus as a vector of counts where dimensions are the words in the vocabulary. The data used are 105 essays and story books also from Adarna House and from the University of the Philippines Integrated School (UPIS). In this work, the TF-IDF representations are decomposed to reduce the sparsity of inputs. Readability assessment was viewed as a clustering task using Latent Semantic Indexing and Content Indexing as primary algorithms of choice for model development. The highest performance obtained is an average overall accuracy of 80.75\% using Content Indexing with Term Frequency. One weakness that we highlight from this approach is that there are no linguistically-motivated features used such as variables leveraging on the semantics and syntactic structure of sentences.
\linebreak

\noindent\textbf{Imperial et al. \shortcite{imperial2019developing}.} Improving the work of Bosque et al. \shortcite{bosquetagatris}, this work explores wider NLP-based features including word n-grams and
character-level n-grams. Similar dataset from Adarna House was used composed of 258 picture books converted to texts. For model development, this study explored on more complex ensemble architecture by combining K-Nearest Neighbors, Random Forest, and Multinomial Naive Bayes through a voting mechanism. Results showed a high performance of 82.2\% using soft voting scheme. One weakness of the study is the use of heavily imbalanced dataset during training.
\linebreak

\noindent\textbf{Imperial and Ong \shortcite{imperial2020exploring}.}
This work pioneered the inclusion of true linguistically-motivated features as prescribed by experts \cite{Macahilig2015} such as lexical features (adapted from Imperial and Ong \shortcite{imperialapplication}) capturing semantics and language model features capturing structure of sentences for a total of 25 predictors overall. For the data, leveled reading materials such as story books and activity books in the first three grade levels of basic education in the Philippines were used. For model development, the study made use of Logistic Regression and Support Vector Machines. Results showed that the inclusion of both lexical and language model features significantly increased the performance of readability assessment models by as high as 25\%-32\% in accuracy (from 33\% to 72.0\% for the top model).
\linebreak

\noindent\textbf{Imperial and Ong \shortcite{imperialong2021}}.
Building from Imperial and Ong \shortcite{imperial2020exploring}, this work explored even deeper linguistic features in Filipino spanning morphological features based on verb inflection and orthographical features based on syllable pattern with a total of 54 predictors overall.  For the data, 265 reading materials from Adarna House (more than the study of Guevarra \shortcite{Guevarra2011}) and DepEd Commons\footnote{https://commons.deped.gov.ph/}, an online and open-source repository of children's resources in Filipino spanning first three grade levels, were used. For model training, the study explored Logistic Regression, Support Vector Machines, and Random Forest. From the 54 linguistic features used, the optimal subset of predictors was the combination of traditional and syllable pattern features obtaining a 66.1\% accuracy using Random Forest. Although relatively lower than in the other previous works, this study empirically showed the learning process of the models as more features are used.
\linebreak

From the notable works listed in line with the development of readability assessment models for Filipino, we specifically chose to interpret the models trained by Imperial and Ong \shortcite{imperialong2021}. We emphasize that this work covers the most number of linguistic features used for any study and is the current state-of-the-art in the context of the Filipino language. Further details of replication and model interpretation are described in the succeeding sections.

\section{Replication Setup}
To interpret relations of model feature to its corresponding output, a trained model is needed first. We obtained such resource directly from the work of Imperial and Ong \shortcite{imperialong2021}.  

\subsection{Corpus}
The study sourced children's books and reading materials from also Adarna House and DepEd Commons. DepEd Commons\footnote{https://commons.deped.gov.ph/} is an online platform launched by the Department of Education (DepEd) of the Philippines. A total of 174 children's fictional books and 91 reading passages were used from both sources respectively. Both datasets have the same granularity of reading levels which are levels 1, 2, and 3 or L1, L2, and L3.

\subsection{Linguistic Features}
The study of Imperial and Ong \shortcite{imperialong2021} extracted 54 different linguistic predictors spanning surface-based, lexical, language model, syllable structure, and morphological features—the most extensive study on the Filipino language to date. These features were also highlighted by past works that should be considered for the readability assessment task \cite{Macahilig2015,Gutierrez2015}. We briefly describe the components of each feature set:\linebreak

\noindent\textbf{Traditional or Surface-Based Features (TRAD).} Word, sentence, phrase counts. Average word length, sentence length, and syllable count per word.\linebreak

\noindent\textbf{Lexical Features (LEX).} Type-token ratio (TTR) and its variations. Noun and verb token ratio. Lexical density. Foreign and compound word density.\linebreak

\noindent\textbf{Language Model Features (LM).} Perplexity scores of documents using language models trained from L1, L2, and L3 books in variations of unigram, bigram, and trigram splits.\linebreak

\noindent\textbf{Syllable Pattern Features (SYLL).} Weighted frequencies of words of a document from the prescribed syllable patterns (\textit{v, cv, vc, vcc, cvc, ccvc, ccvcc, ccvccc}) of the Philippine orthography.\linebreak

\noindent\textbf{Morphological Features (MORPH).} Weighted frequencies of classified words based on morphology, specifically on verb inflection such as focus, aspect, and mood.

\subsection{Model Performance}
We trained the models using the same machine learning algorithms from the study of Imperial and Ong \shortcite{imperialong2021} which were Logistic Regression, Support Vector Machines, and Random Forest. Likewise, we also obtained and set hyperparameter values as used by the study to obtain near similar replication of results. Table~\ref{tab:replicationPerformance} shows the best scoring models using optimal subsets of features. For referencing purposes, the top feature sets for Logistic Regression are TRAD + LEX + SYLL + LM, TRAD + LM for Support Vector Machines, and TRAD + SYLL for Random Forest. These optimized models are highlighted in green in the table.

%
%

\definecolor{top_performing}{RGB}{170, 255, 170}
\definecolor{runner_up}{RGB}{193, 236, 250}
\definecolor{dark_text}{RGB}{105, 144, 91}
\definecolor{light_text}{RGB}{200, 247, 186}

\begin{table}[htbp]\small

    \centering
    \begin{tabular}{|l|c|c|c|c|}
    \hline 
    \bf Model & \bf Acc & \bf Prec & \bf Rec & \bf F1 \\  \hline
    LogReg + All & 0.542 & 0.530 & 0.542 & 0.532  \\ \hline
    
    \rowcolor{light_text} 
    \bf LogReg + Best & \bf 0.576 & \bf 0.544 & \bf 0.576 & \bf 0.561  \\ \hline
    
    SVM + All & 0.492 & 0.481 & 0.492 & 0.485 \\ \hline
    
    \rowcolor{light_text} 
    \bf SVM + Best & \bf 0.525 & \bf 0.524 & \bf 0.525 & \bf 0.524 \\ \hline
    
    RF + All & 0.627 & 0.623 & 0.627 & 0.624 \\ \hline
    
    \rowcolor{light_text}
    \bf RF + Best & \bf 0.661 & \bf 0.651 & \bf 0.661 & \bf 0.640 \\ \hline
    
    \end{tabular}

\caption{Comparison of top-ranked Spearman correlated features against top-ranked features via global interpretation on a Random Forest model.}
\label{tab:replicationPerformance}
\end{table}


\subsection{Top Correlated Features}
In addition to model training, we performed dependence analysis using Spearman correlation to identify if there are predictors from the linguistic features that correlate significantly with the readability levels. Table~\ref{tab:spearmanTable} shows the top 10 Spearman correlated features. From the table, majority of the feature sets are from TRAD and SYLL with a few from LEX and LM. The Type Token Ratio (TTR) emerged as the top negatively correlated feature relative to the task; however, a -0.3379 correlation value does not convey strong relationship by standard. 


\begin{table}[htbp]\small

\centering
\begin{tabular}{|l|l|r|}
\hline 
\bf Feature Set & \bf Predictor & \bf  Spearman's $\rho$  \\  
\hline

LEX   & TTR   & -0.3379 \\ \hline

 & polysyll words & 0.3338  \\ \cline{2-3}
 & average sentence len      & 0.3297  \\ \cline{2-3} 

  \multirow{-3}{*}{TRAD}  & word count                 & 0.2745  \\ \hline

LEX                    & BiLogTTR                    & -0.2723 \\ \hline

 & ccvc density               & 0.2620  \\ \cline{2-3} 
 \multirow{-2}{*}{SYLL} & cvc density                & 0.2300  \\ \hline

LM                     & L1 trigram                 & 0.2247  \\ \hline

  & cvcc density               & 0.2075  \\ \cline{2-3} 
\multirow{-2}{*}{SYLL} & v density                  & 0.1960  \\ \hline

\end{tabular}

\caption{Top 10 ranked Spearman's $\rho$ scores for each linguistic feature.}
\label{tab:spearmanTable}
\end{table}

\section{Global Model Interpretation}
The first part of model interpretation process is the global model interpretability of the readability assessment models trained using Logistic Regression, Support Vector Machines, and Random Forest. Global interpretation is the process of understanding the entire model as a blackbox by looking at the model's \textbf{learned global weights} based on features or possible predictors, in this case, the linguistically-motivated features for readability. Using the ELI5\footnote{https://eli5.readthedocs.io/en/latest/overview.html} package, the interactions of the best combination of the linguistic predictors producing the highest performance from the metrics were obtained for each model. We describe its implications and our findings in this section.

%
%

\begin{figure*}[htbp]
    \centering
    \includegraphics[width=0.98\textwidth]{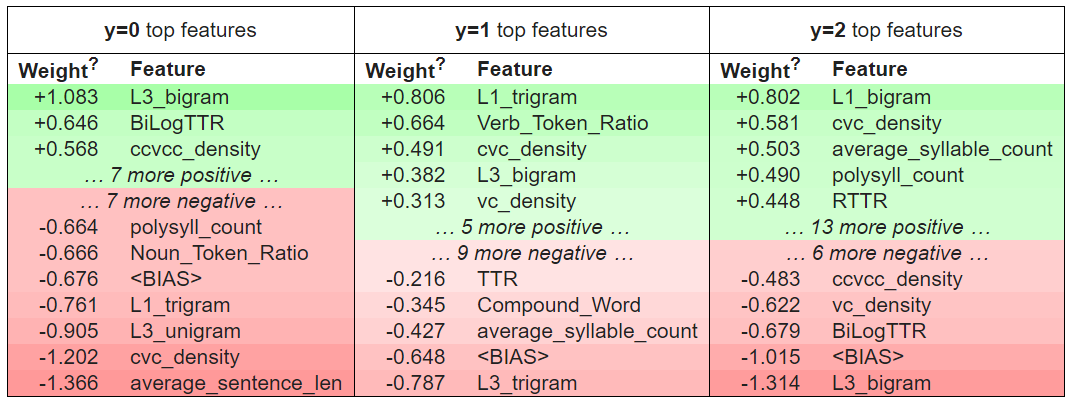}
    \caption{Top 10 highly-important predictors for the Logistic Regression model using all linguistics features from the optimal subset. The green gradient indicates positive relationship while red indicates the opposite.}
    \label{fig:LRWeights}
\end{figure*}

\subsection{Logistic Regression Results}
From the learned weights of the model in Figure~\ref{fig:LRWeights} using the optimal feature subset of the linguistic predictors (TRAD + LEX + SYLL + LM), cross-referencing all features for all grade levels identified that CVC density score and L3 bigram feature were present across all levels. Thus, these two features can be used for readability assessment for Grades 1, 2 and 3. Meanwhile, the top unique features for each grade level are: noun-token ratio and L3 unigram for Grade 1; verb-token ratio, type-token ratio, compound word denstiy, and L3 trigram for Grade 2; and L1 bigram and root type-token ratio for Grade 3. 

Focusing on the top feature with the highest weights for each grade level, for Grade 1 ($y = 0$), the negative weighted average sentence length is the most useful feature for helping the Logistic Regression model classify books as Grade 1. This may mean that \textit{\textbf{as the average length value of sentences decreases, the readability level of texts may increase, proving that readability might not ultimately depend on a direct relationship with sentence length}}. 

On the other hand, for Grade 2 ($y = 1$), the combination of trigrams from the external Grade 1 activity books proved to be effective in helping the model classify story books to its category. For Grade 3 ($y = 2$), the same conclusion can be drawn regarding the inverse relationship of bigrams from external L3 activity books.

%
%

\begin{figure*}[htbp]
    \centering
    \includegraphics[width=0.98\textwidth]{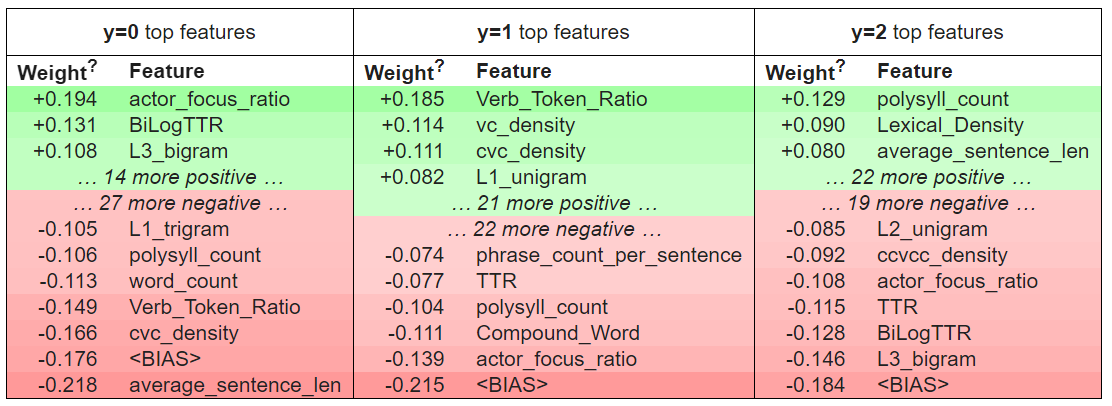}
    \caption{Top 10 highly-important predictors for the Support Vector Machine model using all linguistics features from the optimal subset. The green gradient indicates positive relationship while red indicates the opposite.}
    \label{fig:SVMWeights}
\end{figure*}

\subsection{Support Vector Machine Results}
From the learned weights of the model in Figure~\ref{fig:SVMWeights} using the optimal feature subset of the linguistic predictors (TRAD + LM), cross-referencing all features for all grade levels identified that L1 trigram, L3 unigram, and L3 bigram features were present across all grade levels. Thus, these three features can be used for readability assessment for Grades 1, 2 and 3. While these omnipresent features belong to the LM feature set, the TRAD features used were relatively unique for each grade level. 

For Grade 1 ($y = 0$), the most useful TRAD features are word count observing positive relationship with respect to the readability levels, and average syllable and sentence counts observing negative relationship with respect to the readability levels. For Grade 2 ($y = 1$), the most useful TRAD features are average word length and polysyllable count observing positive and negative relationships with the readability levels respectively. For Grade 3 ($y = 2$), the most useful TRAD features are average sentence length, average syllable count, sentence count, and total polysyllable words all of which observe positive relationships with the readability levels.

%
%

\begin{figure}[htbp]
    \centering
    \includegraphics[width=0.40\textwidth]{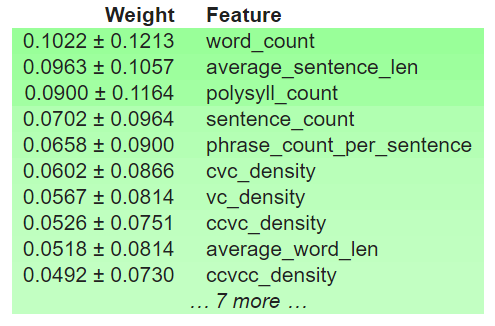}
    \caption{Top 10 highly-important predictors for the Random Forest model using all linguistics features from the optimal subset. All features observe positive relationship with the readability levels.}
    \label{fig:RFWeights}
\end{figure}

\subsection{Random Forest Results}
For the Random Forest model weights as described in Figure~\ref{fig:RFWeights}, the top five features all belong to the TRAD feature set followed by features from SYLL. The simple feature of raw word counts (WC) observing positive relationship with the readability levels emerged as the highest scoring feature. Thus, \textit{\textbf{the higher the word count of a story book, the higher probability of the story book belonging to a higher readability level}}. 

Testing for significant difference\footnote{Two sample $t$ test was conducted over $\alpha = 0.05$.} in the word count of the data from Grade 1 against Grades 2 and 3 obtained $p$ values of 0.004 and 0.014, respectively. Thus, the word count of story books from the Grade 1 dataset is \textit{significantly} different compared to Grades 2 and 3, proving the initial hypothesis.

Interestingly, all of the top TRAD features used by the best Random Forest model are present in traditional, formula-based readability indices. For example, in the formula developed by Villamin \shortcite{Villamin1979}, average sentence length was included; in Guevarra \shortcite{Guevarra2011}, word count, sentence count, and polysyllable words were included; and in Macahilig \shortcite{Macahilig2015}, word count was included. One important inference from this observation is that \textit{\textbf{the use of TRAD features is still practical for readability assessment of Filipino texts, specifically for children's story books}}. However, other deeper linguistic feature sets, such as language model features, can be integrated to these TRAD predictors to produce models capable of achieving better performance for the task.

\subsection{Global Features vs. Correlated Features}
Table~\ref{tab:CorrelationvsInterpretation} compares the performance of running Random Forest models using the top 10 Spearman correlated features from the study of Imperial and Ong \shortcite{imperialong2021} as reported in Table~\ref{tab:spearmanTable} against using the top 10 features reported in Figure~\ref{fig:RFWeights} using global interpretation. From the results, using top weighted features through global interpretation of model obtained much higher performance in contrast to using Spearman features. To be more specific, the increase in scores are 5\% for accuracy, 6.4\% for precision, 7.3\% for recall, and 13.7\% for F1. 

It is also worth mentioning that there are coinciding linguistic features present in both the top Spearman and global interpretation lists such as word count, polysyllable count, and average sentence length. Combining the 15 unique features from both lists and retraining the Random Forest model produces an even higher performance score with 69.8\% in accuracy, $\approx$19\% increase from using global features. With this result, we infer that \textit{\textbf{using top features obtained via global interpretation combined with top correlated Spearman features achieves a much higher performance than using correlated features alone}} as done in previous works.

\begin{table}[htbp]\small

        \begin{tabular}{|l|c|c|c|c|}
        \hline 
        \bf Model & \bf Acc & \bf Prec & \bf Rec & \bf F1 \\  \hline
        RF + Corr & 0.458        & 0.464        & 0.435 & 0.432        \\ \hline
        RF + Global & 0.508        & 0.528         & 0.508 & 0.512       \\ \hline
        \rowcolor{light_text} 
        \bf RF + Combined & \bf 0.698        & \bf 0.682         & \bf 0.624 & \bf 0.649       \\ \hline
        
        \end{tabular}
        \vspace{0.2 cm}

\caption{Comparison of top-ranked Spearman correlated features against top-ranked features via global interpretation on a Random Forest model.}
\label{tab:CorrelationvsInterpretation}
\end{table}

\section{Local Surrogate Interpretation}
The second part of the model interpretation process is the local surrogate interpretability. While global interpretation takes a look at the overall model's learned weights, local surrogate interpretation of models focuses on the local analysis for a particular data point \cite{yang2018global}. We select the best-performing model for this experiment which is the Random Forest model using TRAD + SYLL features as described in Table~\ref{tab:replicationPerformance}. Specifically, the local surrogate interpretation technique used in this experiment is the extraction of \textbf{learned local weight boundaries} used by random tree estimators for correctly classifying different instances from the data of three grade levels. This process was done using the LIME package\footnote{https://github.com/marcotcr/lime} developed in Ribeiro et al \shortcite{ribeiro2016should}.

%
%

\begin{figure*}[htbp]
    \centering
    
    \begin{subfigure}{\linewidth}
        \centering
        $\vcenter{\hbox{\includegraphics[width=0.15\textwidth]{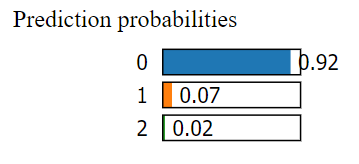}}}$
         $\vcenter{\hbox{\includegraphics[width=0.50\textwidth]{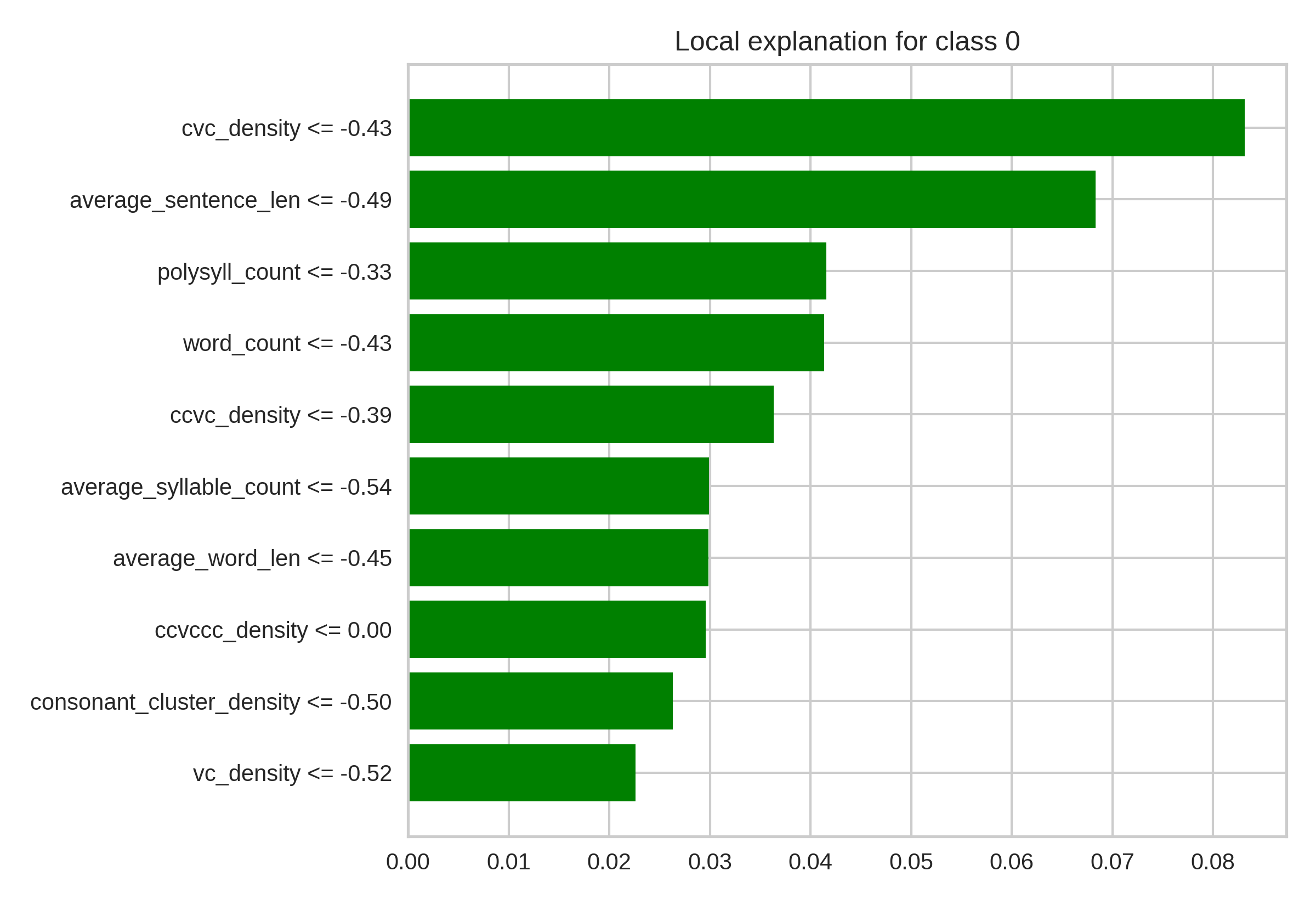}}}$
         
         \subcaption{Grade 1}
    \end{subfigure}
    
    \vspace{0.5cm}
    
    \begin{subfigure}{\linewidth}
        \centering
        $\vcenter{\hbox{\includegraphics[width=0.15\textwidth]{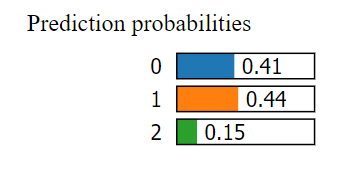}}}$
         $\vcenter{\hbox{\includegraphics[width=0.50\textwidth]{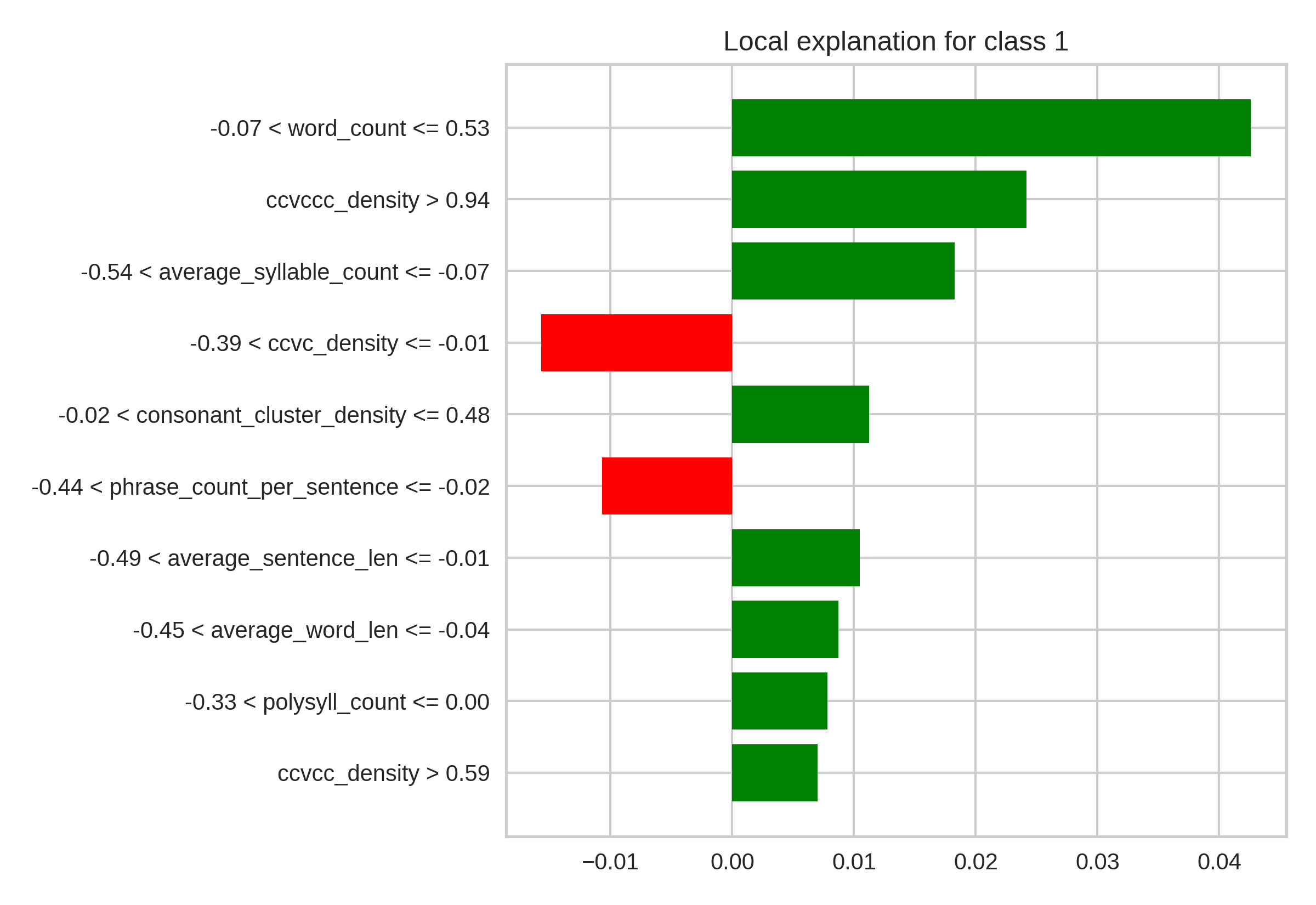}}}$
         
         \subcaption{Grade 2}
    \end{subfigure}
    
    \vspace{0.5cm}
    
    \begin{subfigure}{\linewidth}
        \centering
        $\vcenter{\hbox{\includegraphics[width=0.15\textwidth]{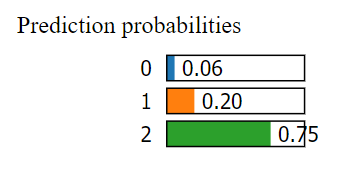}}}$
         $\vcenter{\hbox{\includegraphics[width=0.50\textwidth]{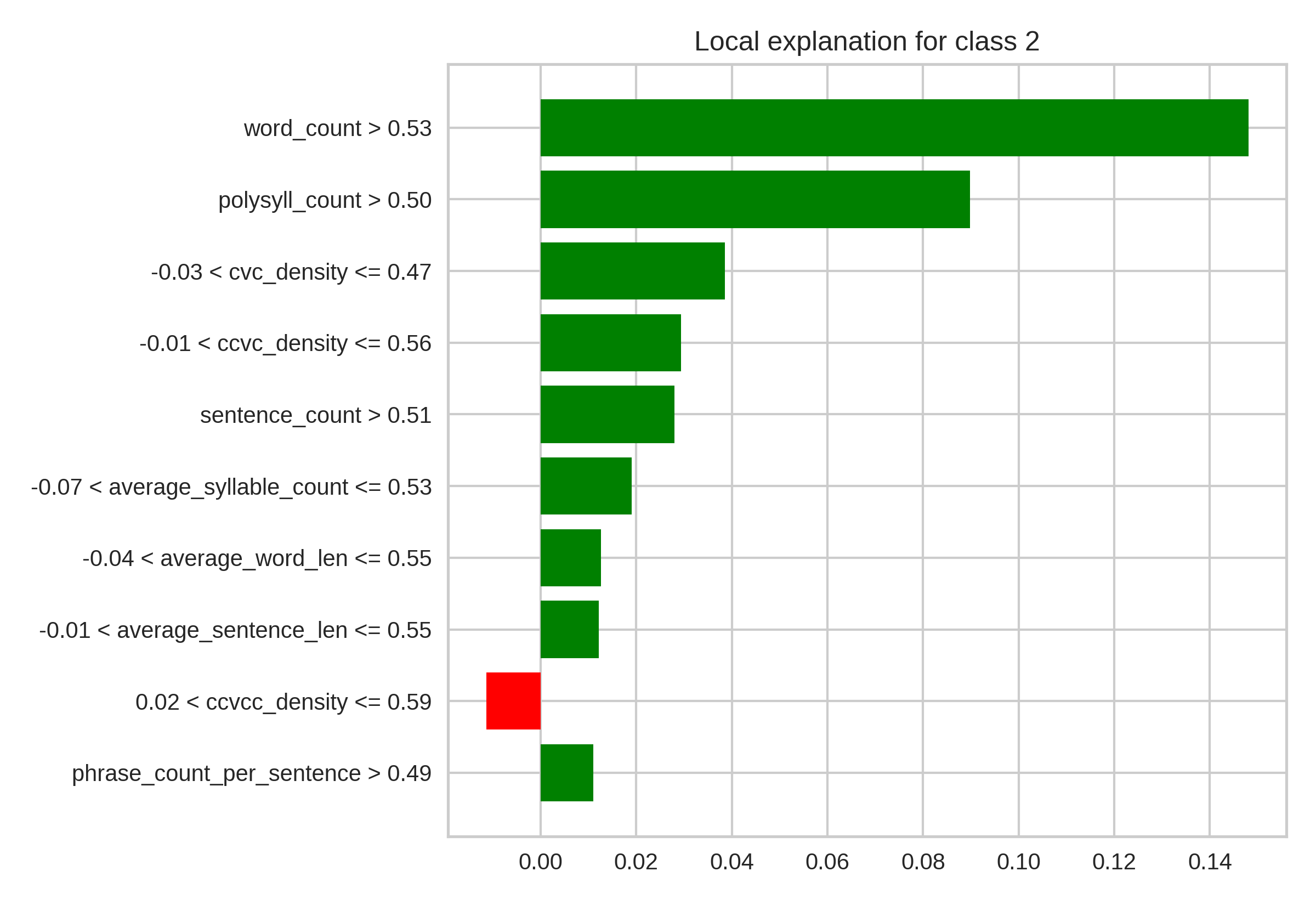}}}$
         
         \subcaption{Grade 3}
    \end{subfigure}
    
    \caption{Local explanations of random test instances for each grade level using the best-performing Random Forest model. Green bars indicates positive relationship while red indicates the opposite.}
    \label{fig:localRFWeights}
\end{figure*}

As seen in Figure~\ref{fig:localRFWeights}, \textbf{the average sentence length is a consistent predictor across all readability levels} used by the best Random Forest model as it is present within the top 10 features. In the case of the test instance for Grade 1, the average sentence length boundary value is $<= -0.43$. Thus, increasing this value would mean that the model may count it towards the other grade levels. For the test instance of Grade 2, the boundary value for average sentence length is $-0.49 < X <= -0.01$, observing a larger range than in the test instance of Grade 1. Thus, if the feature weight of average sentence length falls within this boundary, the model will classify the instance as Grade 2. Similarly, the boundary value of average sentence length for Grade 3 is $-0.04 < X <= 0.55$ wherein if the average weight of a test instance falls within this range, the model will likely label it in favor of Grade 3.

To emphasize, the \textit{connecting} local weight boundaries of linguistic predictors can be traced from Grade 1 to Grade 3 by referencing the values shown in Figure~\ref{fig:localRFWeights}. For example, the boundary value of polsyllable count (frequency of words with more than five syllables)  for Grade 1 is $<= -0.33$, for Grade 2 this becomes $-0.33 < X <= 0.00$, and for Grade 3 this becomes $> 0.50$ indicating that \textit{\textbf{a reading material with a high count of polysyllable words might signal the model to attribute it to a higher readability level}}.

It is important to note that these learned boundaries from the best-performing models are derived only from random instances of the test set. This experiment was performed to determine the conditions that were being used by the learned model to help it discriminate between readability levels. It should also be emphasized that the model will not instantly classify an instance if only one feature or predictor falls within or satisfies the boundary values. Thus, \textit{\textbf{majority of the weights of the top linguistic predictors of a test instance should satisfy the accumulated boundary conditions for it to be classified to a certain category}}.

\section{Conclusion}
Readability assessment is the process of gauging a certain piece of text or reading material in terms of reading difficulty. In this study, we addressed one of the major challenges in readability analysis by exploring feature interpretation using global and local surrogate methods in the context of the Filipino language. Results showed that using global interpretation as a feature selection technique to extract top contributory linguistic features obtained higher performance across all metrics than using the standard correlation method. In addition, combining these two processes would further increase the performance of the model in readability assessment tasks. We also empirically observed the local boundaries of the top-performing model for each linguistic feature. From these boundaries, we learned how perturbing values can directly affect the final readability classification of a certain book or reading material. Future directions of this study include interpretation of complex neural models for readability assessment as there is an increase of patronization in this direction from the research community. Likewise, stronger causality measures such as Bayesian networks can be explored to ground hypotheses on cause and effects of perturbation of local weight boundaries of features. 

\textbf{Acknowledgment} 
The authors would like to thank the anonymous reviewers for their valuable feedback and to Dr. Ani Almario of Adarna House for allowing us to use their children's book dataset for this study. This work is also supported by the DOST National Research Council of the Philippines (NRCP).

\bibliographystyle{acl}
\bibliography{references,anthology}

\end{document}